%% file: acl_latex.tex
\documentclass[11pt]{article}

\usepackage[final]{acl}

\usepackage{times}
\usepackage{latexsym}

\usepackage[T1]{fontenc}
\usepackage[utf8]{inputenc}

\usepackage{microtype}

\usepackage{inconsolata}
\usepackage{booktabs}
\usepackage{amssymb}
\usepackage{amsmath,amsfonts,bm}

\usepackage{graphicx}

\usepackage{enumitem}

\usepackage[table,dvipsnames]{xcolor}
\usepackage{arydshln}
\usepackage{multirow}
\usepackage{makecell}
\usepackage{tikz}
\usepackage{pgfplots}
\usetikzlibrary{patterns}

\pgfplotsset{compat=1.18}  

\newcommand{\eg}{\textit{e.g.}}

\newcommand{\with}{\textit{w/ }}
\newcommand{\without}{\textit{w/o }}

\definecolor{ForestGreen}{rgb}{0.13, 0.55, 0.13} % Defining the ForestGreen color
\definecolor{table_purple}{RGB}{234, 232, 241}
\definecolor{table_green}{RGB}{234, 232, 241}
\definecolor{table_blue}{RGB}{223, 233, 241}
\definecolor{table_yellow}{RGB}{250, 242, 218}

\title{V-CoLA: \underline{V}ision Token \underline{Co}mpression with \underline{L}inear \underline{A}ttention}

\author{
    \textbf{Hao Jiang\textsuperscript{1,*}}\quad
    \textbf{Yiru Mao\textsuperscript{1,*}}\quad
    \\
    \textbf{Tianpeng Bu\textsuperscript{1}}\quad
    \textbf{Hao Zhou\textsuperscript{1}}\quad
    \textbf{Hongtao Duan\textsuperscript{1}}\quad
    \textbf{Wang Jing\textsuperscript{1}}\quad
    \textbf{Bowen Xu\textsuperscript{1}}\quad
    \\
    \textbf{Xin Chen\textsuperscript{1}}\quad
    \textbf{Lulu Hu\textsuperscript{1}}\quad
    \textbf{Bin Yang\textsuperscript{1}}\quad
    \textbf{Yongliang Tao\textsuperscript{1,\textdagger}}\quad
    \textbf{Minying Zhang\textsuperscript{1,\textdagger}}
    \\
    \\
    \textsuperscript{1}Alibaba Cloud Computing, Alibaba Group
    \\
    \small{
        \textsuperscript{*}Equal contribution. \quad
        \textsuperscript{\textdagger}Corresponding author.
    }
    \\
    \small{
        \href{mailto:suanshi.tyl@alibaba-inc.com}{suanshi.tyl@alibaba-inc.com}\quad
        \href{mailto:minying.zmy@alibaba-inc.com}{minying.zmy@alibaba-inc.com}
    }
}

\begin{document}
\maketitle
\input{sections/0_abstract}

\input{sections/1_introduction}
\input{sections/2_analysis}
\input{sections/3_method}
\input{sections/4_experiments}
\input{sections/5_related_work}
\input{sections/6_conclusion}

\input{sections/7_limitations}
\input{sections/8_ethic}
% \section*{Acknowledgments}
% This is Acknowledgments.

\bibliography{custom}

\newpage
\clearpage
\appendix
\input{appendix/1_implementation_parallelism}
\input{appendix/2_more_details}
\input{appendix/3_more_detailed_analysis}

\input{appendix/4_additional_evaluation}
\end{document}

%% file: sections/0_abstract.tex
\begin{abstract}
Vision-language models (VLMs) have demonstrated impressive capabilities but suffer from substantial computational overhead, as vision tokens dominate the input sequence. This motivates vision token compression as a key direction to alleviate the burden. However, with the emergence of hybrid architectures incorporating linear attention (\eg, Qwen3.5), prior methods designed for softmax attention struggle to generalize. Our analysis reveals that both attention- and similarity-based approaches suffer notable performance degradation, underscoring the urgent need for compression methods tailored to this regime. To this end, we propose \textbf{V-CoLA}, an efficient training-free token compression framework specifically designed for linear attention. V-CoLA introduces a novel \textit{uniqueness-aware importance criterion} for identifying critical vision tokens, coupled with an \textit{adaptive token merging strategy} that performs compression. All components are optimized at the implementation level to remain compatible with the chunk-wise parallelism of linear attention, ensuring strong practical value. Extensive experiments across multiple benchmarks demonstrate the superiority of V-CoLA: it achieves 99.5\% of the original performance with only 50.0\% of vision tokens, and over 88.0\% with as few as 12.5\%, while delivering a 1.86$\times$ to 6.15$\times$ prefill speedup. Our code is available at \url{https://github.com/alibaba/EfficientRL}.
\end{abstract}

%% file: sections/1_introduction.tex
\section{Introduction} \label{sec:introduction}
\begin{figure}[!t]
    \centering
    \includegraphics[width=1.0\linewidth]{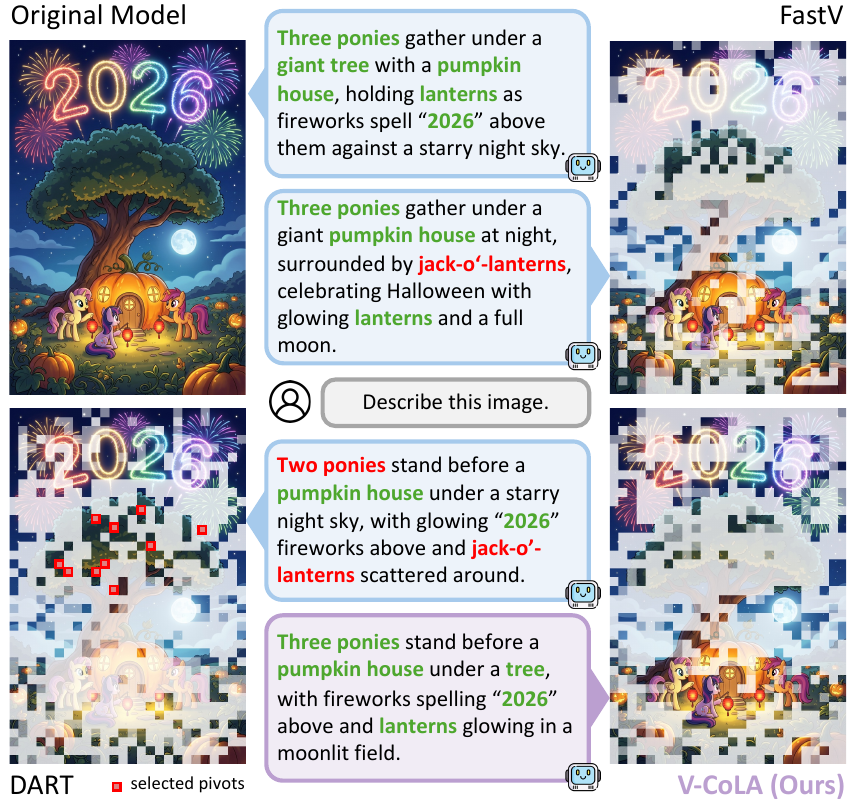}
    \caption{Comparison of token selection. FastV suffers from attention bias, missing the digits at the top of the image. DART selects pivot tokens from only sparse semantic regions, leading to insufficient awareness of some primary objects. Our method successfully identifies all main regions and provides an accurate response.}
    \label{fig:first}
    \vspace{-4mm}
\end{figure}

Vision-language models (VLMs) have demonstrated remarkable performance in multi-modal understanding and reasoning~\cite{qwen3vl,gpt5,internvl35}. Nevertheless, it is widely acknowledged that the number of vision tokens far exceeds that of linguistic tokens~\cite{fastvlm,fastv}, imposing substantial computational and memory burdens. Recent advances in vision token compression~\cite{holitom,dycoke,visionzip,tome} have significantly improved computational efficiency and inference speed by reducing the number of vision tokens. 

However, most of these methods either explicitly rely on softmax-attention scores~\cite{pyramiddrop,lightvlm} or are evaluated solely on softmax-attention models~\cite{dyntok,topv}. With the rise of hybrid architectures incorporating linear attention~\cite{qwen3.5, kimilinear, infinitevl}, the cross-architecture transfer of these methods has become significantly challenging, calling for new compression algorithms tailored to them.

We conduct systematic experiments and find that both attention- and similarity-based compression methods fail to surpass even a random baseline when moving to hybrid architectures, and produce suboptimal results (Figure~\ref{fig:first}). Our theoretical analysis attributes this structural failure to an information-theoretic ceiling imposed by the bounded recurrent state of linear attention, which disrupts both the shallow-layer attention concentration exploited by attention-based methods and the token distinguishability required by similarity-based methods. Yet, it also reveals that linear attention's selective retention itself encodes an intrinsic token-importance signal.

Inspired by this, we explore the intrinsic indicators of token importance inherent in linear attention, and propose \underline{V}ision token \underline{Co}mpression with \underline{L}inear \underline{A}ttention (\textbf{V-CoLA}), a novel efficient token compression framework compatible with hybrid VLMs incorporated with linear attention. Our method mainly evolves two components: (1) an \emph{uniqueness-aware importance criterion} that jointly considers each token’s long-term contribution to the final state and its short-term uniqueness with respect to historical context, enabling the indicating of important tokens while avoiding redundancy, (2) an \emph{adaptive chunk-wise merging strategy} that dynamically adjusts chunk sizes and performs token merging within each chunk, thereby adaptively regulating the local compression ratio and reducing information loss in critical regions. Moreover, we optimize all computations at the implementation level, enabling compatibility with linear attention acceleration algorithms through chunk-wise parallelism.

We conduct comprehensive evaluations using Qwen3.5~\cite{qwen3.5} against state-of-the-art baselines. The experimental results demonstrate that our method significantly outperforms these baselines. Even when preserving 50\% of the tokens after the first layer, V-CoLA achieves 99.5\% of the average performance; when preserving only 12.5\% of the tokens, it still preserves over 88.0\% of the average performance. Benefiting from the implementation-level optimizations in our codebase, V-CoLA achieves a runtime speedup of 1.86$\times$ to 6.15$\times$, while the additional runtime overhead associated with pruning accounts for only 1.6\% of the forward-pass cost.

In summary, our main contributions are:
\begin{itemize}[left=0pt, nosep]
\item We systematically analyze the structural failure of existing vision token compression methods under hybrid architectures with linear attention.
\item We propose V-CoLA, a novel vision token compression framework compatible to linear attention, which achieves superior performance over state-of-the-art baselines.
\item We further optimize V-CoLA for full compatibility with the chunk-wise parallelism of linear attention, achieving significant inference acceleration with marginal computational overhead.
\end{itemize}

%% file: sections/2_analysis.tex
\section{Analysis} \label{sec:analysis}

\subsection{Preliminary} \label{subsec:preliminary}

\input{figs/analysis/comparison_random_baseline}

\paragraph{Softmax Attention.} Given a query $\bm{q}_t$, standard softmax attention computes its output by attending over all preceding key-value pairs:
\begin{equation} \label{eq:softmax_attn}
    \bm{o}_t = \sum_{i=1}^{t} \alpha_{t,i}\, \bm{v}_i, 
    \; \alpha_{t,i} = \frac{\exp(\bm{q}_t^\top \bm{k}_i)}{\sum_{j=1}^{t} \exp(\bm{q}_t^\top \bm{k}_j)},
\end{equation}
where attention weight $\alpha_{t,i}$ provides an explicit, token-level measure of $o_t$'s reliance on $v_i$. While this formulation preserves fine-grained access to every past token, computing the attention map over full sequence incurs quadratic time and memory cost, making it expensive for long contexts.

\paragraph{Linear Attention.} To alleviate the quadratic cost of softmax attention, linear-attention variants replace the softmax kernel with a recurrent state $\bm{S}_t \in \mathbb{R}^{d\times d}$ that compresses the entire history into a fixed-size memory. Among recent advances, Gated DeltaNet~\cite{gateddeltanet} has attracted significant attention and been adopted by popular hybrid VLMs such as Qwen3.5~\cite{qwen3.5} and InfiniteVL~\cite{infinitevl}. It augments the state recurrence with selective writing, erasing, and forgetting operations:
\begin{equation} \label{eq:gated_delta_update}
\begin{gathered}
    \textbf{S}_t = \textbf{S}_{t-1} \bigl(\bm{\alpha}_t (\bm{I} - \beta_t \bm{k}_t \bm{k}_t^\top)\bigr) + \beta_t \bm{v}_t \bm{k}_t^\top,
\end{gathered}
\end{equation}
where $\bm{\alpha}_t \in (0,1)$ is a decay gate, $\beta_t \in (0,1)$ controls the strength of state updating, and $\bm{k}_t$ are $\ell_2$-normalized. The update rule can be interpreted as one step of online gradient descent on a per-token reconstruction objective:
\begin{equation} \label{eq:reconstruction_obj}
    \mathcal{L}_t(\textbf{S})=\tfrac{1}{2}\|\textbf{S}\bm{k}_t-\bm{v}_t\|^2,
\end{equation}
Under this view, the recurrent state $\textbf{S}$ acts as an associative memory that is continually optimized to map each key $\bm{k}_t$ to its corresponding value $\bm{v}_t$. Retrieval at query time is then carried out implicitly through the matrix-vector product $\bm{o}_t = \textbf{S}_t \bm{q}_t,$.

\input{figs/analysis/comparison_vision_attention}

\subsection{Failure of Existing Methods} \label{subsec:existing_fail}
We evaluate the performance of both attention- and similarity-based methods when applied to hybrid architectures across multiple benchmarks. Shown in Figure~\ref{fig:comparison_random_baseline}, they surprisingly exhibit virtually no advantage over the random token pruning baseline. To further investigate the failure mechanisms, we perform an in-depth analysis below.

\begin{figure}[!t]
    \centering
    \includegraphics[width=1.0\linewidth]{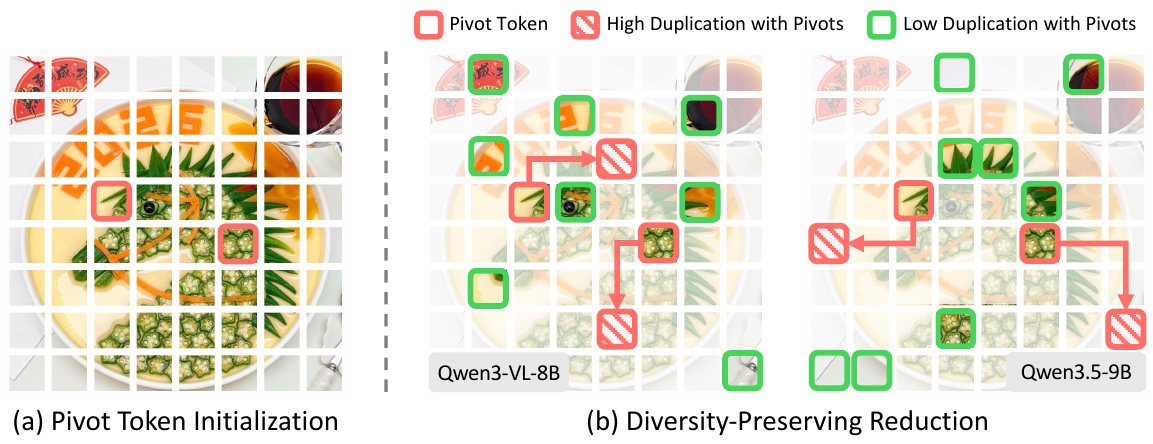}
    \vspace{-6mm}
    \caption{Comparison of similarity-based token selection between Qwen3-VL-8B and Qwen3.5-9B.}
    \label{fig:analysis_similarity_token_selection}
\end{figure}

Figure~\ref{fig:vision_attention_ratio} compares the attention distributions of softmax attention layers at various depths in Qwen3-VL and Qwen3.5. The results illustrate that the vision token attention efficiency of Qwen3-VL declines rapidly in the shallow layers, whereas Qwen3.5 continues to attend to visual content in the intermediate layers and only drops to a low level in the deeper layers. This suggests that leveraging solely on shallow-layer attention may be insufficient to identify all important vision tokens in hybrid architectures. Moreover, we evaluate DART's redundant token selection in Figure~\ref{fig:analysis_similarity_token_selection}. The results show that even with fixed pivot tokens, the redundant tokens identified on Qwen3.5 become markedly less reliable. This suggests that the discriminability of latent feature similarity of hybrid architectures has been fundamentally altered.

The failure of these two representative categories of methods highlights the inherent challenge of transferring previous token compression strategies to hybrid architectures. And these findings indicate that effective token compression for hybrid VLMs requires importance signals that go beyond softmax attention maps and feature similarity, and instead exploit cues intrinsic to the hybrid attention design itself.

\subsection{Information Ceiling of Linear Attention} \label{subsec:analysis_la_ib}
From the associative-memory view, softmax attention keeps every token independently addressable, whereas Gated DeltaNet compresses the entire history into a single fixed-capacity state $\textbf{S}_t \in \mathbb{R}^{d \times d}$ raising an information-preservation problem.

Let $\mathcal{M}=\{(\bm{k}_i,\bm{v}_i)\}_{i=1}^{L}$ be drawn from a joint distribution with per-token entropy $h(\bm{k},\bm{v})$. Treating $\textbf{S}_L$ as a lossy encoder of the sequence and letting $c$ denote the bit-precision of each entry in $\textbf{S}_L$, the data processing inequality~\cite{infotheory} gives $I\bigl(\mathcal{M};\textbf{S}_L\bigr) \leq H(\textbf{S}_L) = c \cdot d^2 = \mathcal{O}(d^{2})$, whereas the source information satisfies $H(\mathcal{M}) = L\cdot h(\bm{k},\bm{v}) = \mathcal{O}(Ld)$. Combining the two yields a lower bound on the information lost by the state:
\begin{equation} \label{eq:info_loss}
\begin{aligned}
    & H\bigl(\mathcal{M}\mid \textbf{S}_L\bigr) = H(\mathcal{M}) - I(\mathcal{M};\textbf{S}_L) \\ 
    & \geq H(\mathcal{M}) - H(\textbf{S}_L) = \Omega(Ld - d^2),
\end{aligned}
\end{equation}
which grows linearly in $L$ once $L>d$. In VLMs, where high-resolution multimodal inputs often yield $L\gg d$, this \emph{information ceiling} becomes particularly pronounced~\cite{arora2024simple, jelassi2024repeat} and carries direct implications for token compression in hybrid architectures. From another perspective, the bounded state compels the gates in Equation~\ref{eq:gated_delta_update} to update selectively, which implies that model's retention preference during state recurrence inherently encodes token importance~\cite{dtp}, providing an intrinsic signal that we exploit in Section~\ref{subsec:la_intrinsic}.

As this qualitative argument relies only on the data-processing inequality and an $L$-independent finite state capacity, it is independent of the specific update rule and extends to bounded-state recurrent backbones in general \cite{mamba2,rwkv}.

%% file: figs/analysis/comparison_random_baseline.tex
\pgfplotsset{compat=1.18}

\begin{figure}[!t]
    \centering
    \hspace*{5mm}  % 向右平移 1cm，负值则向左
    {\tiny
    \begin{tikzpicture} [trim axis left, trim axis right]
        \begin{axis}[
            width=1.1\linewidth,
            height=0.5\linewidth,
            xtick={0,1,2,3,4,5,6,7,8},   % 设置x刻度位置
            xticklabels=\empty,         % 隐藏x刻度标签
            ymin=-18, ymax=12,
            ytick={-10,-5,0,5,10},
            yticklabels={$-10$,$-5$,$0$,$5$,$10$},
            extra y ticks={-15},
            extra y tick labels={$-120$},
            grid=both,
            grid style={line width=.1pt, draw=gray!30},
            major grid style={line width=.2pt,draw=gray!50},
            axis lines=left,
            xmin=0, xmax=8,
            legend style={
                at={(0.3, 0.1)},
                anchor=center,
                draw=none,
                fill=gray!10,
                fill opacity=0.8,
                text opacity=1,
                font=\tiny,
                /tikz/every even column/.append style={column sep=0.3cm},
            },
            legend columns=-1,
            legend cell align={left},
            clip=false,
        ]
            % 绿色背景（上方区域）
            \fill[green!10, opacity=0.5] (axis cs:0,0) rectangle (axis cs:8,12);
            % 红色背景（下方区域）
            \fill[red!10, opacity=0.5] (axis cs:0,-18) rectangle (axis cs:8,0);
            
            % 零线
            \draw[black, thick] (axis cs:0,0) -- (axis cs:8,0);
            
            % FastV 数据点（蓝色圆圈）
            \addplot[only marks, mark=*, mark size=2.5pt, color=blue!60!black] coordinates {
                (0.5, 9.1)
                (1.5, 0.4)
                (2.5, -0.8)
                (3.5, 0.2)
                (4.5, 2.6)
                (5.5, -0.5)
                (6.5, 0.5)
                (7.5, -2.7)
            };
            \addlegendentry{FastV}
            
            % Dart 数据点（橙色三角形）
            \addplot[only marks, mark=triangle*, mark size=3pt, color=orange!90!red] coordinates {
                (0.5, -15.1)
                (1.5, -1.1)
                (2.5, -1.5)
                (3.5, -0.2)
                (4.5, -7.6)
                (5.5, -0.4)
                (6.5, 0.9)
                (7.5, 0.2)
            };
            \addlegendentry{Dart}
            
            % 数据集标签
            \node at (axis cs:0.5, 2.5) {MME};
            \node at (axis cs:1.5, 2.5) {MMB};
            \node at (axis cs:2.5, 2.5) {GQA};
            \node at (axis cs:3.5, 2.5) {SQA};
            \node at (axis cs:4.5, -2.5) {T-VQA};
            \node at (axis cs:5.5, 2.5) {POPE};
            \node at (axis cs:6.5, -2.5) {VizWiz};
            \node at (axis cs:7.5, 2.5) {MMStar};
            
            % 区域文字标注
            \node[green!50!black, font=\tiny] at (axis cs:6.0, 9.5) {Outperforms Random Baseline};
            \node[red!70!black, font=\tiny] at (axis cs:6.0, -16) {Underperforms Random Baseline};
            
            % 断轴标记
            \draw[black, thin] (axis cs:-0.05,-13.5) -- (axis cs:0.05,-12.5);
            \draw[black, thin] (axis cs:-0.05,-13.0) -- (axis cs:0.05,-12.0);
            
        \end{axis}
    \end{tikzpicture}}
    \vspace{-2mm}
    \caption{FastV (attention-based) \textit{vs.} DART (similarity-based) \textit{vs.} random baseline on Qwen3.5-9B.}
    \label{fig:comparison_random_baseline}
    \vspace{-4mm}
\end{figure}

%% file: figs/analysis/comparison_vision_attention.tex
\begin{figure}[!t]
\centering
{\tiny
\begin{tikzpicture}
\begin{axis}[
    width=1.05\linewidth,
    height=0.515\linewidth,
    ybar,
    bar width=12pt,
    ylabel={Vision Attention Ratio (\%)},
    symbolic x coords={layer4, layer12, layer24},
    xtick=data,
    ymin=0, ymax=10,
    ytick={0,2,4,6,8,10},
    nodes near coords,
    nodes near coords style={font=\tiny},
    legend image code/.code={
        \draw [#1] (0cm,-0.1cm) rectangle (0.2cm,0.1cm);
    },
    legend style={
        at={(0.98,0.98)},
        anchor=north east,
        font=\tiny,
        draw=none,
        fill=white,
        fill opacity=0.8,
        text opacity=1,
    },
    axis lines=box,
    axis line style={line width=0.3pt, draw=gray!60},
    tick style={line width=0.3pt, draw=gray},
    xtick pos=bottom,
    grid=none,
    line width=0.3pt,
    enlarge x limits=0.3,
]

\addplot[
    draw=purple!50!white,
    fill=purple!40!white,
] coordinates {
    (layer4, 3.41)
    (layer12, 1.39)
    (layer24, 0.82)
};
\addlegendentry{Qwen3-VL-8B}

\addplot[
    draw=purple!70!black,
    fill=purple!70!black,
] coordinates {
    (layer4, 8.76)
    (layer12, 8.54)
    (layer24, 0.97)
};
\addlegendentry{Qwen3.5-9B}

% 折线1：连接 Qwen3-VL-8B 柱顶（虚线，向左偏移）
\addplot[
    sharp plot,
    draw=purple!50!white,
    line width=0.8pt,
    dashed,
    mark=*,
    mark size=1.5pt,
    mark options={fill=purple!50!white, draw=purple!50!white, solid},
    forget plot,
    xshift=-7pt,
    every node near coord/.style={opacity=0},
] coordinates {
    (layer4, 3.41)
    (layer12, 1.39)
    (layer24, 0.82)
};

% 折线2：连接 Qwen3.5-9B 柱顶（虚线，向右偏移）
\addplot[
    sharp plot,
    draw=purple!70!black,
    line width=0.8pt,
    dashed,
    mark=*,
    mark size=1.5pt,
    mark options={fill=purple!70!black, draw=purple!70!black, solid},
    forget plot,
    xshift=7pt,
    every node near coord/.style={opacity=0},
] coordinates {
    (layer4, 8.76)
    (layer12, 8.54)
    (layer24, 0.97)
};

\end{axis}
\end{tikzpicture}}
\vspace{-2mm}
\caption{Average proportion of attention scores allocated to vision tokens across different layers in Qwen3-VL-8B and Qwen3.5-9B.}
\label{fig:vision_attention_ratio}
\end{figure}

%% file: sections/3_method.tex
\section{Method} \label{sec:method}
\begin{figure*}[!t]
\begin{center}
\centerline{\includegraphics[width=2.1\columnwidth]{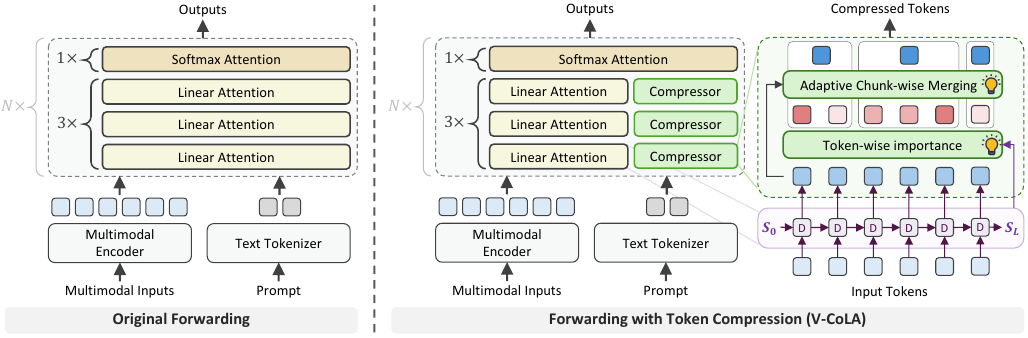}}
\vspace{-2mm}
\caption{Overview of our method. (left) Original hybrid architectures. (right) Forwarding with V-CoLA, equipped with token-wise importance criterion and adaptive chunk-wise token merging.}
\label{fig:overview}
\end{center}
\vspace{-8mm}
\end{figure*}
In Section \ref{sec:analysis}, we analyze the failure of transferring typical attention-based and similarity-based methods to identify token importance into hybrid architectures, and the information ceiling of linear attention. Experimental results show different behavioral patterns in the state recurrence and pose challenges for token importance estimation and compression methods. To address this, shown in Figure~\ref{fig:overview}, we propose a novel training-free token compression framework. Specifically, we first uncover the inherent capability of linear attention to indicate token importance in Section~\ref{subsec:la_intrinsic}.  Building upon this insight, we further introduce a uniqueness-aware token importance in Section~\ref{subsec:token_importance}, and develop an adaptive chunk-wise token merging strategy in Section~\ref{subsec:token_merging}.

\subsection{Linear Attention as an Intrinsic Indicator} \label{subsec:la_intrinsic}
The analysis in Section~\ref{subsec:analysis_la_ib} reveals that the update process of the bounded recurrent state inherently encodes token importance, which motivates us to derive it directly from the state recurrence.

Although pioneering work~\cite{dtp} has attempted to quantify token importance through the "magnitude of state update", such a strategy considers only a single dimension of the recurrence and lacks a measure of long-term information retention. According to Equation~\ref{eq:reconstruction_obj}, the objective of state recurrence is to write the paired key-value information into the state such that it can be fully retrieved. We adopt this reconstruction error as a straightforward criterion, and quantify the retention rate of visual information written into the final state of the model across varying vision token lengths and model depths. We also record the Pearson correlation coefficient of the retention rate rankings between adjacent layers.

As shown in Figure~\ref{fig:retention_rate}, we have several key observations: (1) The retention rate is relatively high in shallow layers and progressively declines with depth, dropping below 20\% in the deeper layers. This indicates that visual information is heavily involved in shallow-layer reasoning, while becoming increasingly marginalized in deeper layers. (2) The retention rate also decreases as the sequence length grows, which aligns with the information ceiling discussed in Section~\ref{subsec:analysis_la_ib}. Given the limited state size, this suggests that the recurrent state is already under pressure in retaining information, forcing the model to preserve tokens in a selective manner. (3) The Pearson correlation of retention rate rankings between adjacent layers remains high in shallow layers and declines in deeper layers alongside the fading of visual information. This implies that during the visual reasoning stage, different layers exhibit consistent preferences in selecting critical tokens.

The above observations offer important insights: the state updates of linear attention inherently possess a capability for indicating token importance, and we can leverage the selective retention behavior of linear attention in the shallow layers to filter out important or redundant tokens.

\input{figs/analysis/retention_rate_v4}

\subsection{Uniqueness-aware Token Importance} \label{subsec:token_importance}
As discussed in Section~\ref{subsec:la_intrinsic}, the reconstruction objective of linear attention provides a natural lens for measuring token importance: a low reconstruction error indicates that the token's information is faithfully preserved in the state. Building on this insight, follow-up work~\cite{gatedkalmanet} further posits that state updates ought to account for both current and historical information, proposing a more comprehensive objective:
\begin{equation} 
\begin{aligned}
    \mathcal{L}_t(\textbf{S})=\lambda\cdot\|S\|^2_F + \sum^t_{i=1}\eta_i\cdot\|\textbf{S}\bm{k}_i-\bm{v}_i\|^2.
\end{aligned}
\end{equation}
This suggests that an important token should be accurately encoded in the state $\textbf{S}$, while the degree of reconstruction error \textbf{reflects whether the token has contributed to the final state}. Inspired by this, we propose a token importance criterion from the perspective of memory reconstruction:
\begin{equation} \label{eq:r_imp}
\begin{aligned}
    \mathcal{R}^{\text{imp}}_t=1-\psi(\textbf{S}_L\bm{k}_t, \bm{v}_t),\quad t\in[1,L],
\end{aligned}
\end{equation}
where $S_L$ is the final state of the sequence with length $L$, and $\psi(\mathbf{a},\mathbf{b})=\left(1-\cos(\mathbf{a},\mathbf{b})\right)/2\in[0,1]$ denotes the normalized cosine distance.

However, filtering based solely on token-wise importance suffers from a bias toward salient regions, thereby causing the model to select a large number of redundant tokens~\cite{dart}. Moreover, focusing on the final state may cause certain important long-term tokens to be overlooked, as their contributions may be attenuated by the accumulated gated decaying in Equation~\ref{eq:gated_delta_update}, or chooses to write information and pass it directly to the next layer for processing, rather than continuously propagating along the sequence to $\textbf{S}_L$.

To overcome this, we introduce token’s short-term contribution to the state as a complement to Equation~\ref{eq:r_imp}. In the recurrent update of the state, each token writes its information into the state and optimize the objective in Equation~\ref{eq:reconstruction_obj}. As a result, when subsequent redundant tokens further update the state, the gradient of the optimization objective decreases, which in turn suppresses the magnitude of their influence on the state. Therefore, the degree of state change can be used to measure redundancy and serve as an indicator of uniqueness. To assess this change more robustly, we use a set of pseudo queries $Q^*_t\in\mathbb{R}^{w \times d}$ as probes for querying instead of comparing state after/before updating, and introduce following metric:
\begin{equation} \label{eq:r_uni}
\begin{aligned}
    \mathcal{R}^{\text{uni}}_t=\psi(\textbf{S}_t\textbf{Q}^{*\top}_t, \textbf{S}_{t-1}\textbf{Q}^{*\top}_t),\quad t\in[1,L],
\end{aligned}
\end{equation}
where we choose $Q^*_t=Q_{[t+w_1:t+w_2)}$ as the nearest queries from token $t$, which will be affected in recent state recurrent updates, with $w_1\leq 0$ and $w_2>0$ denoting the signed left and right boundary offsets, respectively. Each of the $w=w_2-w_1$ probes reads one value vector from the state before and after token $t$ is absorbed, and then we average $\psi$ over the probes.

Overall, we define the final importance score in a weighted form:
\begin{equation} \label{eq:r_final}
\begin{aligned}
    \mathcal{R}_t = (1-\lambda) \cdot \mathcal{R}^{\text{imp}}_t + \lambda \cdot \mathcal{R}^{\text{uni}}_t,
\end{aligned}
\end{equation}
where $\lambda$ denotes the weight of $\mathcal{R}^{\text{uni}}_t$. As both terms are $\psi$-based distances over value-space readouts, $\mathcal{R}_t$ itself remains a normalized measure in $[0,1]$, consistently interpreted within the value space.

Notably, we integrate the importance computation into the chunk-wise parallelism of linear attention, incurring minimal additional runtime overhead. The details are provided in the Appendix~\ref{ap_sec:implementation_parallelism}.

\subsection{Adaptive Chunk-wise Token Merging}  \label{subsec:token_merging}
Previous approaches primarily remain tokens with top-$k$ importance scores~\cite{topv,gsearch,fastv}.  Recent analyses~\cite{clusterandaggregate} indicate that, compared with token pruning, token aggregation exhibits more robust performance. Inspired by this, various token aggregation strategies have been proposed, \eg, clustering-based aggregation~\cite{fastvid,holitom} and bipartite matching aggregation~\cite{folder}. However, these methods do not consider distribution of importance, therefore fail to adaptively allocate compression rate across varying semantic regions. 

Based on these, we propose an adaptive token merging strategy that automatically partitions tokens into importance-aligned chunks according to their importance distribution density, and performs token merging within each chunk. Given token sequence $o_t$, remain rate $\gamma$ and importance $\mathcal{R}_{t}$, we partition the token sequence into $L'=L\times\gamma$ chunks such that the sum of importance scores within each chunk is close. Specifically, we define chunks as $\mathcal{C}_i=o_{[t_i: t_{i+1}]}, 1=t_1 < \cdots < t_{L'}=L$, the objective can be formulated as:
\begin{equation}
\begin{aligned}
    \{\mathcal{C}^*_i\}^{L'}_{i=1}=\arg\min_{\{\mathcal{C}_i\}}\sum^{L'}_{i=1}\left|\bar{\mathcal{R}}-\sum^{t_{i+1}}_{t=t_i}\mathcal{R}_t\right|,
\end{aligned}
\end{equation}
where $\bar{\mathcal{R}}=\frac{1}{L'}\sum^{L}_{t=1}\mathcal{R}_t$ is expected importance mass per chunk. In this way, regions with dense importance are assigned finer-grained chunks, while less informative regions are grouped into coarser chunks, yielding an importance-aware partition of the token sequence. Then, we perform token merging independently within each chunk and get compressed token sequence $o'_i$ by:
\begin{equation}
\begin{aligned}
    o'_i=\sum^{t_{i+1}}_{t=t_i}w_t o_t,\; w_t=\frac{\exp(\mathcal{R}_t/\tau)}{\sum^{t_{i+1}}_{k=t_i}\exp(\mathcal{R}_k/\tau)},
\end{aligned}
\end{equation}
where $\tau$ is a temperature parameter that controls the smoothness of token merging, which is set as $1.0$ in our experiments. After that, $o'_t$ is further propagated through the subsequent layers. 

In Appendix~\ref{ap_sec:implementation_parallelism}, we present optimized implementation of the proposed algorithm, including efficient chunk partitioning and token merging via matrix parallel operators, which achieves negligible runtime latency in practice.

\subsection{Early Exiting of Vision Tokens} \label{subsec:overall_pipeline}
\input{tables/main_comparison}
As suggested by previous work~\cite{hidrop} that deeper layers tend to perform vision-independent reasoning, we observe a similar pattern in the hybrid architecture: the proportion of important visual tokens sharply decreases in later layers, as shown in Figure~\ref{fig:retention_rate}. To further validate this observation, we conduct a perplexity (PPL) analysis by systematically removing all vision tokens at each layer and measuring the resulting PPL relative to the full-token baseline:
\input{figs/analysis/early_exit_curve}
where the PPL drops sharply and plateaus after layer-24 of the model, corroborating the importance pattern observed that vision tokens become increasingly negligible in deep layers. We also report the MME benchmark results that discarding vision tokens beyond different layer depths:
\begin{center}
{\small
\begin{tabular}{ccccc}
\toprule
\rowcolor{gray!15}
\textbf{Exit Layer} & \textbf{1} & \textbf{12} & \textbf{24} & \textbf{32 (N/A)} \\
MME     & 967.3 & 2169.5 & 2401.6  & 2398.2 \\
MMB     & 22.6   & 80.4   & 85.2   & 85.6 \\
SQA     & 83.7   & 90.6   & 92.3   & 92.2 \\
\bottomrule
\end{tabular}}
% \vspace{1mm}
\end{center}
which demonstrates that the model can fully preserve its multimodal reasoning performance even without vision tokens in the deep layers. Based on these, we propose a vision token early-exit mechanism that discards all vision tokens in the deep layers of the model, which further reducing the computational burden.

%% file: figs/analysis/retention_rate_v4.tex
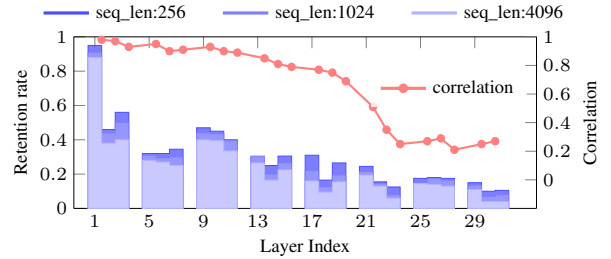
\begin{figure}[!t]
\centering
{\scriptsize
\begin{tikzpicture} [trim axis left, trim axis right]
\begin{axis}[
    width=1.0\linewidth,
    height=0.5\linewidth,
    xlabel={Layer Index},
    ylabel={Retention rate},
    xmin=1, xmax=32,
    ymin=0, ymax=1,
    enlarge x limits=0.05,
    xtick={1,5,9,13,17,21,25,29},
    ytick={0.0,0.2,0.4,0.6,0.8,1.0},
    legend style={
        font=\scriptsize,
        draw=none,
        fill=white,
        fill opacity=0.8,
        text opacity=1,
        at={(0.5,1.03)},
        anchor=south,
        legend columns=-1,
        /tikz/every even column/.append style={column sep=0.4cm},
    },
    legend image post style={line width=1.2pt},
    axis line style={line width=0.3pt},
    tick style={line width=0.3pt},
    grid=none,
    line width=0.3pt,
    axis y line*=left,
]

%==================== seq_len:256 ====================
\addplot[draw=blue!70!white, fill=blue!60!white, fill opacity=0.85]
coordinates {(0.5,0) (0.5,0.95) (1.5,0.95) (1.5,0.46) (2.5,0.46) (2.5,0.56) (3.5,0.56) (3.5,0)} \closedcycle;
\addlegendentry{seq\_len:256}
\addplot[draw=blue!70!white, fill=blue!60!white, fill opacity=0.85, forget plot]
coordinates {(4.5,0) (4.5,0.32) (5.5,0.32) (5.5,0.32) (6.5,0.32) (6.5,0.345) (7.5,0.345) (7.5,0)} \closedcycle;
\addplot[draw=blue!70!white, fill=blue!60!white, fill opacity=0.85, forget plot]
coordinates {(8.5,0) (8.5,0.47) (9.5,0.47) (9.5,0.45) (10.5,0.45) (10.5,0.40) (11.5,0.40) (11.5,0)} \closedcycle;
\addplot[draw=blue!70!white, fill=blue!60!white, fill opacity=0.85, forget plot]
coordinates {(12.5,0) (12.5,0.30) (13.5,0.30) (13.5,0.25) (14.5,0.25) (14.5,0.305) (15.5,0.305) (15.5,0)} \closedcycle;
\addplot[draw=blue!70!white, fill=blue!60!white, fill opacity=0.85, forget plot]
coordinates {(16.5,0) (16.5,0.31) (17.5,0.31) (17.5,0.165) (18.5,0.165) (18.5,0.265) (19.5,0.265) (19.5,0)} \closedcycle;
\addplot[draw=blue!70!white, fill=blue!60!white, fill opacity=0.85, forget plot]
coordinates {(20.5,0) (20.5,0.245) (21.5,0.245) (21.5,0.155) (22.5,0.155) (22.5,0.125) (23.5,0.125) (23.5,0)} \closedcycle;
\addplot[draw=blue!70!white, fill=blue!60!white, fill opacity=0.85, forget plot]
coordinates {(24.5,0) (24.5,0.175) (25.5,0.175) (25.5,0.18) (26.5,0.18) (26.5,0.175) (27.5,0.175) (27.5,0)} \closedcycle;
\addplot[draw=blue!70!white, fill=blue!60!white, fill opacity=0.85, forget plot]
coordinates {(28.5,0) (28.5,0.15) (29.5,0.15) (29.5,0.10) (30.5,0.10) (30.5,0.105) (31.5,0.105) (31.5,0)} \closedcycle;

%==================== seq_len:1024 ====================
\addplot[draw=blue!50!white, fill=blue!40!white, fill opacity=0.9]
coordinates {(0.5,0) (0.5,0.91) (1.5,0.91) (1.5,0.44) (2.5,0.44) (2.5,0.50) (3.5,0.50) (3.5,0)} \closedcycle;
\addlegendentry{seq\_len:1024}
\addplot[draw=blue!50!white, fill=blue!40!white, fill opacity=0.9, forget plot]
coordinates {(4.5,0) (4.5,0.31) (5.5,0.31) (5.5,0.295) (6.5,0.295) (6.5,0.30) (7.5,0.30) (7.5,0)} \closedcycle;
\addplot[draw=blue!50!white, fill=blue!40!white, fill opacity=0.9, forget plot]
coordinates {(8.5,0) (8.5,0.44) (9.5,0.44) (9.5,0.435) (10.5,0.435) (10.5,0.395) (11.5,0.395) (11.5,0)} \closedcycle;
\addplot[draw=blue!50!white, fill=blue!40!white, fill opacity=0.9, forget plot]
coordinates {(12.5,0) (12.5,0.305) (13.5,0.305) (13.5,0.20) (14.5,0.20) (14.5,0.265) (15.5,0.265) (15.5,0)} \closedcycle;
\addplot[draw=blue!50!white, fill=blue!40!white, fill opacity=0.9, forget plot]
coordinates {(16.5,0) (16.5,0.22) (17.5,0.22) (17.5,0.125) (18.5,0.125) (18.5,0.195) (19.5,0.195) (19.5,0)} \closedcycle;
\addplot[draw=blue!50!white, fill=blue!40!white, fill opacity=0.9, forget plot]
coordinates {(20.5,0) (20.5,0.215) (21.5,0.215) (21.5,0.145) (22.5,0.145) (22.5,0.08) (23.5,0.08) (23.5,0)} \closedcycle;
\addplot[draw=blue!50!white, fill=blue!40!white, fill opacity=0.9, forget plot]
coordinates {(24.5,0) (24.5,0.15) (25.5,0.15) (25.5,0.175) (26.5,0.175) (26.5,0.165) (27.5,0.165) (27.5,0)} \closedcycle;
\addplot[draw=blue!50!white, fill=blue!40!white, fill opacity=0.9, forget plot]
coordinates {(28.5,0) (28.5,0.135) (29.5,0.135) (29.5,0.07) (30.5,0.07) (30.5,0.08) (31.5,0.08) (31.5,0)} \closedcycle;

%==================== seq_len:4096 ====================
\addplot[draw=blue!30!white, fill=blue!20!white, fill opacity=0.95]
coordinates {(0.5,0) (0.5,0.88) (1.5,0.88) (1.5,0.38) (2.5,0.38) (2.5,0.40) (3.5,0.40) (3.5,0)} \closedcycle;
\addlegendentry{seq\_len:4096}
\addplot[draw=blue!30!white, fill=blue!20!white, fill opacity=0.95, forget plot]
coordinates {(4.5,0) (4.5,0.28) (5.5,0.28) (5.5,0.27) (6.5,0.27) (6.5,0.25) (7.5,0.25) (7.5,0)} \closedcycle;
\addplot[draw=blue!30!white, fill=blue!20!white, fill opacity=0.95, forget plot]
coordinates {(8.5,0) (8.5,0.40) (9.5,0.40) (9.5,0.395) (10.5,0.395) (10.5,0.335) (11.5,0.335) (11.5,0)} \closedcycle;
\addplot[draw=blue!30!white, fill=blue!20!white, fill opacity=0.95, forget plot]
coordinates {(12.5,0) (12.5,0.265) (13.5,0.265) (13.5,0.165) (14.5,0.165) (14.5,0.225) (15.5,0.225) (15.5,0)} \closedcycle;
\addplot[draw=blue!30!white, fill=blue!20!white, fill opacity=0.95, forget plot]
coordinates {(16.5,0) (16.5,0.16) (17.5,0.16) (17.5,0.095) (18.5,0.095) (18.5,0.155) (19.5,0.155) (19.5,0)} \closedcycle;
\addplot[draw=blue!30!white, fill=blue!20!white, fill opacity=0.95, forget plot]
coordinates {(20.5,0) (20.5,0.195) (21.5,0.195) (21.5,0.13) (22.5,0.13) (22.5,0.06) (23.5,0.06) (23.5,0)} \closedcycle;
\addplot[draw=blue!30!white, fill=blue!20!white, fill opacity=0.95, forget plot]
coordinates {(24.5,0) (24.5,0.145) (25.5,0.145) (25.5,0.14) (26.5,0.14) (26.5,0.13) (27.5,0.13) (27.5,0)} \closedcycle;
\addplot[draw=blue!30!white, fill=blue!20!white, fill opacity=0.95, forget plot]
coordinates {(28.5,0) (28.5,0.11) (29.5,0.11) (29.5,0.04) (30.5,0.04) (30.5,0.04) (31.5,0.04) (31.5,0)} \closedcycle;

\end{axis}

% 右侧第二个坐标轴用于折线图（未修改）
\begin{axis}[
    width=1.0\linewidth,
    height=0.5\linewidth,
    xmin=1, xmax=32,
    ymin=-0.2, ymax=1.0,
    xtick=\empty,
    axis y line*=right,
    axis x line=none,
    enlarge x limits=0.05,
    ylabel={Correlation},
    ylabel near ticks,
    ytick={0.0,0.2,0.4,0.6,0.8,1.0},
    legend style={
        font=\scriptsize,
        draw=none,
        fill=white,
        fill opacity=0.8,
        text opacity=1,
        at={(0.98,0.7)},
        anchor=east,
    },
    legend image post style={line width=1.2pt},
    axis line style={line width=0.3pt},
    tick style={line width=0.3pt},
    line width=0.3pt,
]
\addplot[
    color=red!50,
    mark=*,
    mark size=1.2pt,
    line width=0.8pt,
] coordinates {
    (1.5,0.98) (2.5,0.97) (3.5,0.93) (5.5,0.95) (6.5,0.90) (7.5,0.91) (9.5,0.93) (10.5,0.90) (11.5,0.89) (13.5,0.85) (14.5,0.81) (15.5,0.79) (17.5,0.77) (18.5,0.75) (19.5,0.69) 
    (21.5,0.51) (22.5,0.35) (23.5,0.25)(25.5,0.27) (26.5,0.29) (27.5,0.21)  (29.5,0.25) (30.5,0.27) 
};
\addlegendentry{correlation}

\end{axis}
\end{tikzpicture}}
\vspace{-2mm}
\caption{Layer-wise retention rate (blue) and importance correlation (red) on Qwen3.5-9B.}
\label{fig:retention_rate}
\vspace{-4mm}
\end{figure}

%% file: tables/main_comparison.tex
\begin{table*}[t]
    % \vspace{-3mm}
    \caption{We compare our method with state-of-the-art approaches using Qwen3.5-9B. The best results are bold.}
    \vspace{-2mm}
    \label{tab:main_comparison}
    \scriptsize
    \centering
    \resizebox{\textwidth}{!}{
    \begin{tabular}{l cccccccc c}
        \toprule
        \textbf{Method} &
        \rotatebox{90}{\textbf{MME}} &
        \rotatebox{90}{\textbf{MMB}} &
        \rotatebox{90}{\textbf{GQA}} &  
        \rotatebox{90}{\textbf{SQA}} &  
        \rotatebox{90}{\textbf{T-VQA}} & 
        \rotatebox{90}{\textbf{POPE}} &  
        \rotatebox{90}{\textbf{VizWiz}} & 
        \rotatebox{90}{\textbf{MMStar}} & 
        \rotatebox{90}{\textbf{Avg(\%)}} \\
        \midrule
        \multicolumn{10}{c}{\cellcolor{gray!15}\textit{Upper Bound, 880 Tokens \textbf{(100\%)}}} \\
        Qwen3.5-9B                    & 2398.2 & 85.6 & 61.1 & 92.2 & 83.2 & 89.9 & 69.2 & 49.3 & 100.0 \\
        \midrule
        \multicolumn{10}{c}{\cellcolor{gray!15}\textit{Remain 440 Tokens in Average \textcolor{ForestGreen}{\textbf{($\downarrow$ 50.0\%)}}}} \\
        FastV~\cite{fastv}            & 2351.6 & 84.8 & 60.4 & 92.1 & 79.5 & 88.8 & 67.6 & 45.3 & 97.5  \\
        SparseVLM~\cite{sparsevlm}    & 2333.1 & 84.7 & 60.1 & 92.2 & 80.1 & 88.7 & 67.5 & 45.6 & 97.5  \\
        DART~\cite{dart}              & 2220.4 & 83.3 & 59.6 & 91.7 & 69.3 & 88.9 & 67.0 & 48.3 & 95.5  \\
        VisionZip~\cite{visionzip}    & 2376.5 & 84.7 & 60.7 & 92.6 & 80.6 & 89.4 & 68.0 & 48.9 & 99.0  \\
        DTP~\cite{dtp}                & 2197.2 & 81.6 & 58.7 & 91.0 & 71.5 & 87.3 & 67.6 & 45.0 & 94.2  \\
        V-CoLA (Ours)                 & \cellcolor{table_purple}{\textbf{2398.5}} & \cellcolor{table_purple}{\textbf{85.0}} & \cellcolor{table_purple}{\textbf{61.0}} & \cellcolor{table_purple}{\textbf{92.7}} & \cellcolor{table_purple}{\textbf{80.6}} & \cellcolor{table_purple}{\textbf{89.8}} & \cellcolor{table_purple}{\textbf{68.3}} & \cellcolor{table_purple}{\textbf{49.6}} & \cellcolor{table_purple}{\textbf{99.5}}  \\
        \midrule
        \multicolumn{10}{c}{\cellcolor{gray!15}\textit{Remain 220 Tokens in Average \textcolor{ForestGreen}{\textbf{($\downarrow$ 75.0\%)}}}} \\
        FastV~\cite{fastv}            & 2219.8 & 83.7 & 57.4 & 90.4 & 58.9 & 85.7 & 64.4 & 42.4 & 90.9  \\
        SparseVLM~\cite{sparsevlm}    & 2252.8 & 83.5 & 55.1 & 89.9 & 61.8 & 85.4 & 63.5 & 43.7 & 91.1 \\
        DART~\cite{dart}              & 2025.8 & 79.7 & 55.6 & 89.9 & 52.3 & 86.2 & 65.5 & 43.3 & 88.4  \\
        VisionZip~\cite{visionzip}    & 2247.1 & 83.6 & 59.1 & 92.1 & \textbf{65.7} & 88.0 & 66.1 & 47.0 & 94.5  \\
        DTP~\cite{dtp}                & 1975.3 & 76.2 & 53.9 & 87.9 & 58.6 & 79.7 & 66.7 & 39.6 & 86.3  \\
        V-CoLA (Ours)                 & \cellcolor{table_purple}{\textbf{2268.7}} & \cellcolor{table_purple}{\textbf{84.4}} & \cellcolor{table_purple}{\textbf{60.1}} & \cellcolor{table_purple}{\textbf{92.4}} & \cellcolor{table_purple}{63.4} & \cellcolor{table_purple}{\textbf{88.9}} & \cellcolor{table_purple}{\textbf{66.7}} & \cellcolor{table_purple}{\textbf{47.2}} & \cellcolor{table_purple}{\textbf{94.8}}  \\
        \midrule
        \multicolumn{10}{c}{\cellcolor{gray!15}\textit{Remain 110 Tokens in Average \textcolor{ForestGreen}{\textbf{($\downarrow$ 87.5\%)}}}} \\
        FastV~\cite{fastv}            & 1387.6 & 56.5 & 43.0 & 84.2 & 14.6 & 61.4 & 57.0 & 28.3 & 63.9  \\
        SparseVLM~\cite{sparsevlm}    & 1622.4 & 68.1 & 43.7 & 85.3 & 16.5 & 69.5 & 57.1 & 33.0 & 69.7  \\
        DART~\cite{dart}              & 1792.4 & 71.3 & 47.6 & 87.8 & 38.2 & 81.4 & 61.7 & 37.9 & 79.2  \\
        VisionZip~\cite{visionzip}    & \textbf{2036.0} & 77.3 & 56.7 & 90.8 & 45.7 & 84.1 & 65.3 & 40.3 & 86.4  \\
        DTP~\cite{dtp}                & 1721.4 & 64.7 & 47.6 & 85.2 & 45.9 & 68.2 & 63.2 & 32.4 & 75.7  \\
        V-CoLA (Ours)                 & \cellcolor{table_purple}{2012.0} & \cellcolor{table_purple}{\textbf{77.9}} & \cellcolor{table_purple}{\textbf{57.5}} & \cellcolor{table_purple}{\textbf{90.9}} & \cellcolor{table_purple}{\textbf{46.0}} & \cellcolor{table_purple}{\textbf{88.3}} & \cellcolor{table_purple}{\textbf{68.3}} & \cellcolor{table_purple}{\textbf{41.4}} & \cellcolor{table_purple}{\textbf{88.0}}  \\
        \bottomrule
    \end{tabular}
    }
    % \vspace{-3mm}
\end{table*}

%% file: figs/analysis/early_exit_curve.tex
\pgfplotsset{compat=1.18}

\begin{tikzpicture} [trim axis left, trim axis right]
\begin{axis}[
    width=1.1\columnwidth,
    height=0.50\columnwidth,
    xlabel={\scriptsize Exiting Layer},
    ylabel={\scriptsize PPL ratio},
    xlabel style={
        yshift=4pt,
    },
    ylabel style={
        at={(axis description cs:-0.01,0.6)},
        anchor=south west,
    },
    xmin=0, xmax=33,
    ymin=0.9, ymax=3,
    xtick={1,4,8,12,16,20,24,28,32},
    ytick={1.00,2.00},
    tick label style={font=\scriptsize},
    xticklabel style={yshift=2pt},
    yticklabel style={xshift=2pt},
    grid=both,
    grid style={dashed, gray!30},
    tick align=outside,
]

% Shaded area under curve down to baseline
\addplot[
    draw=none,
    fill=red!15,
    forget plot,
] coordinates {
    (1,0.98) (1,2.87) (2,2.86) (3,2.86) (4,2.88) (5,2.88) (6,2.88) (7,2.88)
    (8,2.46) (9,2.46) (10,2.46) (11,2.46)
    (12,2.56) (13,2.56) (14,2.56) (15,2.56)
    (16,1.88) (17,1.88) (18,1.88) (19,1.88)
    (20,1.09) (21,1.09) (22,1.09) (23,1.09)
    (24,1.02) (25,1.02) (26,1.02) (27,1.01)
    (28,1.00) (29,1.00) (30,1.00) (31,1.00) (32,0.98)
    (32,0.98)
} \closedcycle;

% Main line: PPL Ratio
\addplot[
    color=red!80!black,
    thick,
    mark=*,
    mark size=1.2pt,
] coordinates {
    (1,2.87) (2,2.86) (3,2.86) (4,2.88) (5,2.88) (6,2.88) (7,2.88)
    (8,2.46) (9,2.46) (10,2.46) (11,2.46)
    (12,2.56) (13,2.56) (14,2.56) (15,2.56)
    (16,1.88) (17,1.88) (18,1.88) (19,1.88)
    (20,1.09) (21,1.09) (22,1.09) (23,1.09)
    (24,1.02) (25,1.02) (26,1.02) (27,1.01)
    (28,1.00) (29,1.00) (30,1.00) (31,1.00) (32,0.98)
};

% Baseline ratio = 1
\addplot[
    color=gray,
    dashed,
    thick,
] coordinates {(0,1) (33,1)};

% Baseline label directly on the line
\node[gray, font=\scriptsize, anchor=south west] at (axis cs:0.6,1.0) {baseline (ratio=1)};
\node[red!80!black, font=\scriptsize, anchor=south west] at (axis cs:0.6,2.4) {early exit};

% Vertical dotted line at L24
\draw[red, dotted, thick] (axis cs:24,0.9) -- (axis cs:24,2.95);
\node[red, font=\scriptsize] at (axis cs:26.5,1.35) {ppl ratio $\approx$ 1 @ $L_{24}$};
\end{axis}
\end{tikzpicture}

%% file: sections/4_experiments.tex
\section{Experiments} \label{sec:experiments}
\subsection{Setup}
\input{tables/runtime_generation}
\paragraph{Models and Benchmarks.} Our experiments are mainly conducted on Qwen3.5-9B~\cite{qwen3.5}, which is a popular hybrid model family with linear attention. To validate the generalizability of the proposed method, we also include Qwen3.5-27B and InfiniteVL~\cite{infinitevl} in additional experiments.
All experiments are performed on NVIDIA A100 GPUs.

We conduct experiments on comprehensive benchmarks: MME~\cite{mme}, MMB~\cite{mmbench}, GQA~\cite{gqa}, ScienceQA~\cite{sqa}, TextVQA~\cite{textvqa}, POPE~\cite{pope}, VizWiz~\cite{vizwiz} and MMStar~\cite{mmstar}. We employ LMMs-Eval~\cite{lmmseval} for benchmark evaluation.

\paragraph{Hyperparameters.} In our main experiments, importance-based token compression is applied by default at the end of layer-1 (whereas applied at the first softmax-attention layer for attention-score-based methods), and the early exit in V-CoLA is set to layer-24. To ensure a fair comparison, we align the average token remain rates across different methods in our experiments. We set distance function $\psi(\cdot,\cdot)$ as cosine distance. The window parameters $(w_1, w_2)$ in Equation~\ref{eq:r_uni} are set as $(0, 4)$ and the weight $\lambda$ in Equation~\ref{eq:r_final} is set as $0.1$.

\input{tables/runtime_component}

\subsection{Main Results}
\paragraph{Comparison with State-of-the-art Methods.} We select several advanced and representative methods for comparison: (1) FastV~\cite{fastv}, (2) SparseVLM~\cite{sparsevlm}, (3) DART~\cite{dart}, (4) VisionZip~\cite{visionzip}, (5) DTP~\cite{dtp}. Notably, DTP is also specifically designed for the Linear Attention architecture. As shown in Table~\ref{tab:main_comparison}, our method consistently outperforms the compared approaches across different benchmarks under various compression rates.

\paragraph{Efficiency of V-CoLA.} To demonstrate the practical performance of our method in real-world applications, we compare the actual multimodal inference latency under varying remain rates. As shown in Table~\ref{tab:runtime_gen}, V-CoLA achieves consistent reductions in prefill latency compared to the original model, attaining a speedup from $1.86 \times$ to $6.15 \times$. Furthermore, we report the runtime of the main components in Table~\ref{tab:runtime_comp}. Notably, our implementation of the Extended Gated DeltaRule maintains comparable latency to the original operator under single-query inference, while substantially outperforming it under 4-query inference, achieving a 3.85$\times$ speedup. Owing to our efficient implementation, the Adaptive Token Merging incurs only 0.86ms of overhead, accounting for approximately 0.1\% of the total model prefill cost. This highlights the deployment friendliness of V-CoLA in practical applications. We also provide FLOPs analysis in Appendix~\ref{ap_sec:flops_analysis} and an end-to-end latency breakdown in Appendix~\ref{ap_sec:additional_eval}.

\subsection{More Analysis}
\input{tables/ablation_lambda}
\paragraph{Ablation of $\lambda$.} The parameter $\lambda$ controls the weight of $\mathcal{R}^{\text{uni}}_t$ in Equation~\ref{eq:r_final}. We compare different $\lambda$ in Table~\ref{tab:ablation_lambda}, the results show that $\lambda=0.1$ produces better performance.

\input{tables/ablation_w}
\paragraph{Ablation of $(w_1, w_2)$.} We also compare different setting of window parameters $(w_1,w_2)$ in Equation~\ref{eq:r_uni}, where we find that expanding a single pseudo query into a window of length 4 enhances robustness and improves performance, with forward window selection proving more advantageous than backward selection. However, further increasing the number of queries beyond this point yields no additional performance gains.

\paragraph{Hyperparameter sensitivity.} Tables~\ref{tab:ablation_lambda} and~\ref{tab:ablation_w} vary smoothly rather than sharply, where $\lambda\in[0.05,0.10]$ performs comparably and only an extreme value degrades notably, and setting $(w_1, w_2)$ to $(0,4)$ and $(-4,4)$ yields closely matching results. Moreover, we adopt an identical configuration for Qwen3.5-9B, Qwen3.5-27B and InfiniteVL in Tables~\ref{tab:main_comparison} and~\ref{tab:ext_comparison} and obtain the best results on all three, indicating that our default setting transfers across model scales and architecture families without per-model tuning.

\paragraph{Ablation of main components.} We additionally ablate two key components: adaptive token merging and early exit of vision tokens. As shown in Table~\ref{tab:ablation_main}, removing either component leads to performance drops. This confirms the complementary roles of both components: adaptive merging provides a more comprehensive way to aggregate token information than hard pruning, while early exit enables us to reallocate the redundant token budget in deeper layers toward shallow-layer visual understanding. Notably, by tuning the temperature coefficient, our merging strategy can also emulate hard pruning, affording users greater flexibility for customization.

\input{tables/ablation_main}
\paragraph{Generalizability of V-CoLA.} As a supplementary analysis, we extend comparisons to different model scales (Qwen3.5-27B) and model families (InfiniteVL). As shown in Table~\ref{tab:ext_comparison}, V-CoLA consistently achieves leading performance across models and benchmarks, demonstrating the strong generalization of our method. Evaluations on long-form multi-turn dialogue and small-object recognition are further provided in Appendix~\ref{ap_sec:additional_eval}.

Together with Table~\ref{tab:main_comparison}, our evaluation thus covers representative hybrid linear-attention VLMs across two model scales and two architecture families. Beyond these models, both criteria remain well defined for recurrent backbones with compatible matrix-state readouts, including updates of the general form $\textbf{S}_t=\textbf{A}_t\textbf{S}_{t-1}+\textbf{B}_t\bm{v}_t\bm{k}_t^\top$, and extending the evaluation to such backboned models is a natural direction for future work.
\input{tables/ext_comparison}

%% file: tables/runtime_generation.tex
\begin{table}[!ht]
    \centering
    % \vspace{-2mm}
    \caption{Comparison of wall-clock runtime (ms) of the language model prefilling across various vision token numbers and different remain rates.}
    \vspace{-2mm}
    \resizebox{1.0\columnwidth}{!}{
        \begin{tabular}{lcccc}
        \toprule
        \textbf{\# V-Tokens} & \textbf{100\%} & \textbf{50.0\%} & \textbf{25.0\%} & \textbf{12.5\%} \\
        \midrule
        2K      & 269    & 145 (1.86 $\times$)   & 88 (3.06 $\times$)   & 61 (4.41 $\times$) \\
        4K      & 511    & 273 (1.87 $\times$)   & 153 (3.34 $\times$)  & 94 (5.44 $\times$)\\
        8K      & 1034   & 542 (1.91 $\times$)   & 289 (3.58 $\times$)  & 168 (6.15 $\times$) \\
        \bottomrule 
        \end{tabular}
    }
    \label{tab:runtime_gen}
    % \vspace{-2mm}
\end{table}

%% file: tables/runtime_component.tex
\begin{table}[!t]
    \centering
    % \vspace{-2mm}
    \caption{Comparison of wall-clock runtime across different components.}
    \vspace{-2mm}
    \resizebox{1.0\columnwidth}{!}{
        \begin{tabular}{lcc}
        \toprule
        \textbf{Component}  & \textbf{\# Tokens} & \textbf{Runtime (ms)} \\
        \midrule
        Softmax Attention                              & 8K $\times$ 1 & 68.5 \\
        \hdashline
        \multirow{2}{*}{Gated DeltaRule}               & 8K $\times$ 1 & 34.1 \\
                                                     ~ & 8K $\times$ 4 & \underline{135.1} \\
        \hdashline
        \multirow{2}{*}{Extended Gated DeltaRule}      & 8K $\times$ 1 & 34.2 \\
                                                     ~ & \cellcolor{table_purple}{8K $\times$ 4} & \cellcolor{table_purple}{\textbf{35.1} (3.85$\times$)} \\
        \hdashline
        Adaptive Token Merging                         & \cellcolor{table_purple}{8K $\times$ 1} & \cellcolor{table_purple}{\textbf{0.86}} \\
        \bottomrule
        \end{tabular}
    }
    \label{tab:runtime_comp}
    % \vspace{-2mm}
\end{table}

%% file: tables/ablation_lambda.tex
\begin{table}[!t]
    \centering
    % \vspace{-2mm}
    \caption{Ablation of parameter $\lambda$ in $\mathcal{R}_t$ calculation.}
    \vspace{-2mm}
    \resizebox{1.0\columnwidth}{!}{
        \begin{tabular}{lcccc}
        \toprule
            \textbf{$\lambda$}                          & \textbf{MME} & \textbf{SQA} & \textbf{POPE} & \textbf{MMStar} \\
            \midrule
            $0.00$ (only $\mathcal{R}^{\text{imp}}_t$)  & 2360.5 & 92.3 & 89.5 & 48.9 \\ 
            $0.05$                                      & 2388.0 & 92.5 & 89.6 & 49.6 \\
            $0.10$                                      & \cellcolor{table_purple}{\textbf{2398.5}} & \cellcolor{table_purple}{\textbf{92.7}} & \cellcolor{table_purple}{\textbf{89.8}} & \cellcolor{table_purple}{\textbf{49.8}} \\
            $0.90$                                      & 2333.1 & 91.7 & 87.3 & 48.0 \\
        \bottomrule
        \end{tabular}
    }
    \label{tab:ablation_lambda}
    % \vspace{-2mm}
\end{table}

%% file: tables/ablation_w.tex
\begin{table}[!t]
    \centering
    \caption{Ablation of window parameters $(w_1, w_2)$.}
    \vspace{-2mm}
    \resizebox{0.85\columnwidth}{!}{
        \begin{tabular}{ccccc}
        \toprule
            \textbf{$(w_1, w_2)$}        & \textbf{MME} & \textbf{SQA} & \textbf{POPE} & \textbf{MMStar} \\
            \midrule
            $(0, 1)$                       & 2387.6 & 92.5 & 89.3 & 49.1 \\ 
            $(0, 4)$                       & \cellcolor{table_purple}{\textbf{2398.5}} & \cellcolor{table_purple}{\textbf{92.7}} & \cellcolor{table_purple}{89.8} & \cellcolor{table_purple}{\textbf{49.8}} \\
            $(-4, 0)$                      & 2376.4 & 92.4 & 87.9 & 49.3 \\
            $(-4, 4)$                      & 2395.2 & 92.6 & \textbf{89.9} & 49.5 \\
        \bottomrule
        \end{tabular}
    }
    \label{tab:ablation_w}
\end{table}

%% file: tables/ablation_main.tex
\begin{table}[!t]
    \centering
    \caption{Ablation of main components.} 
    \vspace{-2mm}
    \resizebox{1.0\columnwidth}{!}{
        \begin{tabular}{lcccc}
        \toprule
            \textbf{Method}                          & \textbf{MME} & \textbf{SQA} & \textbf{POPE} & \textbf{MMStar} \\
            \midrule
            \multicolumn{5}{c}{\cellcolor{gray!15}\textit{\textcolor{ForestGreen}{\textbf{$\downarrow$ 75.0\%}} Vision Tokens}} \\
            V-CoLA                   & \textbf{2268.7}  & \textbf{92.4}  & \textbf{88.9}  & \textbf{47.2} \\
            - w/o Ada. Merging       & 2197.4 & 91.5 & 88.4 & 44.3 \\
            - w/o Early Exit         & 2235.1 & 91.9 & 88.2 & 45.6 \\
            \midrule
            \multicolumn{5}{c}{\cellcolor{gray!15}\textit{\textcolor{ForestGreen}{\textbf{$\downarrow$ 87.5\%}} Vision Tokens}} \\
            V-CoLA                   & \textbf{2012.0}  & \textbf{90.9}  & \textbf{88.3}  & \textbf{41.4} \\
            - w/o Ada. Merging       & 1913.4 & 89.7 & 86.2 & 36.5 \\
            - w/o Early Exit         & 1970.3 & 89.4 & 87.6 & 38.1 \\
        \bottomrule
        \end{tabular}
    }
    \label{tab:ablation_main}
\end{table}

%% file: tables/ext_comparison.tex
\begin{table}[t]
    \caption{We extend comparison with state-of-the-art approaches to include Qwen3.5-27B and InfiniteVL, with fixed \textcolor{ForestGreen}{\textbf{$\downarrow$ 75.0\%}} token compression.}
    \vspace{-2mm}
    \label{tab:ext_comparison}
    \footnotesize
    \centering
    \resizebox{\columnwidth}{!}{
    \begin{tabular}{l cccc}
        \toprule
        \textbf{Method} &
        \textbf{MME} &
        \textbf{SQA} &
        \textbf{POPE} &  
        \textbf{MMStar} \\
        \midrule
        \cellcolor{gray!15}{Qwen3.5-27B}          & \cellcolor{gray!15}{2517.4} & \cellcolor{gray!15}{97.0} & \cellcolor{gray!15}{90.5} & \cellcolor{gray!15}{58.1} \\
        - DART                 & 2173.0 & 91.9 & 82.4 & 44.5 \\
        - VisionZip            & 2405.2 & 96.0 & 88.4 & 53.3 \\
        - DTP                  & 1798.2 & 89.2 & 70.2 & 43.1 \\
        - V-CoLA (Ours)        & \cellcolor{table_purple}{\textbf{2414.5}} & \cellcolor{table_purple}{\textbf{96.1}} & \cellcolor{table_purple}{\textbf{88.8}} & \cellcolor{table_purple}{\textbf{53.7}} \\
        \midrule
        \cellcolor{gray!15}{InfiniteVL}          & \cellcolor{gray!15}{1998.9} & \cellcolor{gray!15}{86.0} & \cellcolor{gray!15}{87.9} & \cellcolor{gray!15}{53.7} \\
        - DART                 & 1885.5 & 84.7 & 85.0 & 49.0 \\
        - VisionZip            & 1883.8 & 84.8 & \textbf{86.8} & 48.4 \\
        - DTP                  & 1663.8 & 83.0 & 82.0 & 44.1 \\
        - V-CoLA (Ours)        & \cellcolor{table_purple}{\textbf{1893.1}} & \cellcolor{table_purple}{\textbf{85.8}} & \cellcolor{table_purple}{86.3} & \cellcolor{table_purple}{\textbf{49.3}} \\
        \bottomrule
    \end{tabular}
    }
\end{table}

%% file: sections/5_related_work.tex
\section{Related Work} \label{sec:related_work}

\subsection{Visual Token Compression for MLLMs}
Since visual tokens often dominate the input sequence in MLLMs \cite{fastv,visionzip}, token compression has emerged as a key direction for reducing computational cost. Existing methods fall into two categories. \textit{Attention-based} methods prune tokens via softmax attention scores, either in a single step \cite{fastv}, progressively across layers \cite{pyramiddrop}, or with text-guided sparsification and feature recycling \cite{sparsevlm,visionzip,ficoco}. \textit{Similarity-based} methods provide attention-free alternatives through bipartite matching \cite{tome,auroracap}, feature-spatial similarity \cite{topv}, or diversity-aware selection \cite{dart}. However, the former relies on softmax attention scores, which are sparsely available and exhibit different distributional properties in hybrid linear-attention architectures. The latter assumes that hidden representations preserve reliable similarity structure, an assumption challenged by the state-recurrent updates in linear attention, where outputs are queried from a shared compressed state. While pioneering work~\cite{dtp} have explored token compression for mamba-based architectures, it merely leverage intermediate output for importance ranking with suboptimal performance.

A parallel line of work reduces tokens by \textit{grouping} rather than scoring, \eg, the Visual-Word Tokenizer \cite{vwt} merges patches into visual words according to intra- and inter-image statistics. Such grouping is performed at the encoder stage from static patch-space statistics, and is therefore complementary to our importance signal derived from state recurrence inside the backbone.

\subsection{Linear Attention Architectures}
Beyond the simple recurrent accumulation of early linear attention \cite{linearattn}, recent work introduce more expressive updating paradigms. DeltaNet \cite{deltanet} erases existing associations along the current key direction before writing new ones, and Gated DeltaNet \cite{gateddeltanet} further adds per-dimension decay gates for selective forgetting. More broadly, the same principle underlies state-space and recurrent alternatives such as Mamba-2 \cite{mamba2} and RWKV \cite{rwkv}, which likewise compress the history into a bounded state under data-dependent gating. These designs underpin hybrid architectures such as Qwen3.5 \cite{qwen3.5} and Kimi-Linear~\cite{kimilinear}, which interleave linear recurrence with softmax attention, and have been extended to multimodal settings \cite{visualrwkv,matvlm,infinitevl}. Nevertheless, token compression tailored to linear-attention VLMs remains largely unexplored, motivating architecture-aware importance estimation.

%% file: sections/6_conclusion.tex
\section{Conclusion} \label{sec:conclusion}
We present V-CoLA, a training-free vision token compression framework tailored to hybrid VLMs with linear attention. Motivated by the observation that previous token compression method fail to transfer to hybrid architectures, V-CoLA estimates token importance from the intrinsic dynamics of linear attention, and further incorporates an adaptive chunk-wise merging strategy and an early-exit mechanism for vision tokens. With implementation-level optimizations compatible with chunk-wise parallelism, V-CoLA delivers state-of-the-art performance across diverse benchmarks while substantially improving inference efficiency.

%% file: sections/7_limitations.tex
\section{Limitations}
While V-CoLA demonstrates strong empirical performance and efficiency gains, we acknowledge several limitations of the current study, which also point to promising directions for future work.

\paragraph{Architectural coverage.}
Our evaluation primarily focuses on hybrid VLMs that interleave linear-attention and softmax-attention layers, such as Qwen3.5 and InfiniteVL. While this family represents one of the most widely adopted designs in recent linear-attention VLMs, other architectural variants, including purely Mamba-based or fully linear-attention VLMs, are not yet covered in our experiments. Extending V-CoLA to such architectures may require revisiting the interaction between our uniqueness-aware importance criterion and the absence of periodic softmax-attention layers, which we leave for future exploration.

\paragraph{Scope of linear-attention formulations.}
Our analysis and method design are mainly grounded in the Gated DeltaNet formulation, which underlies several recent hybrid VLMs. Although the reconstruction-based perspective we adopt is fairly general and may extend to other recurrent state update rules (\eg, other gated linear attention variants), a more systematic study across a broader range of linear-attention formulations would be valuable for assessing the generality of the proposed criterion.

%% file: sections/8_ethic.tex
\section{Ethical Considerations}
This work focuses on improving the inference efficiency of vision-language models through a training-free token compression framework. We conduct no user studies or user-facing deployment, and no human-subject experiments are involved. All experiments are performed on publicly available benchmarks and pretrained models, following their intended research use. These resources contain no intended personally identifiable or sensitive information. Since V-CoLA is training-free and operates on top of existing VLMs, it does not introduce additional biases beyond those already present in the underlying models.

%% file: appendix/1_implementation_parallelism.tex
\section{V-CoLA Implementation Parallelism} \label{ap_sec:implementation_parallelism}

\begin{figure*}[!t]
\begin{center}
\centerline{\includegraphics[width=2.1\columnwidth]{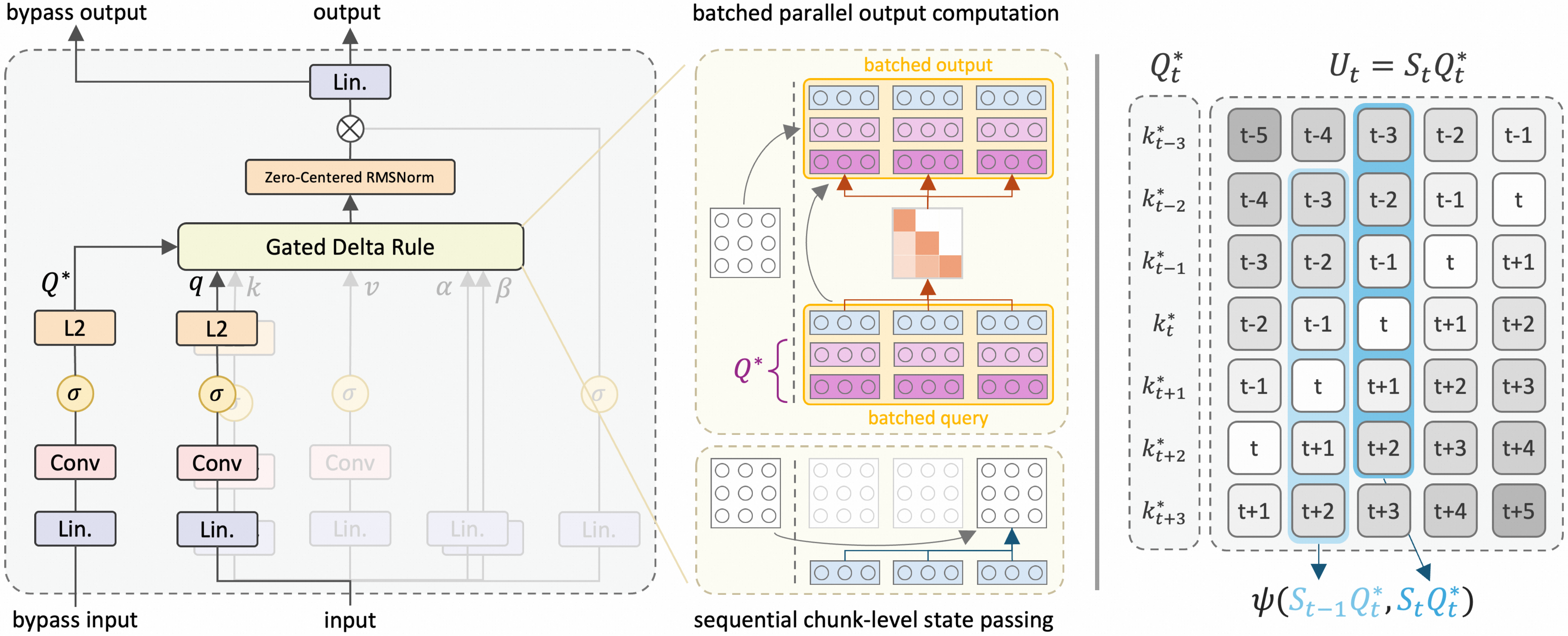}}
\caption{Parallelism algorithm of token uniqueness. (left) We extend Gated DeltaNet with bypass input and batched queries, it enables injecting additional queries when computing output. (right) We use shifted keys as pseudo queries, and calculate token uniqueness on output matrix $U_t$.}
\label{fig:parallel_state_delta}
\end{center}
\end{figure*}

Making token compression compatible with existing inference acceleration techniques has long remained highly challenging~\cite{lightvlm, topv, glimpseprune}. In this section, we present a parallel implementation of the main components introduced above, aiming to maximize compatibility with linear-attention acceleration algorithms~\cite{kimilinear, gateddeltanet} while incurring minimal additional overhead.

\paragraph{Token importance $\mathcal{R}^{\text{imp}}_t$}. Since the computation of $\mathcal{R}^{\text{imp}}_t$ depends only on the original input and the final state $S_L$, it is naturally compatible with various forms of chunk-wise parallelism while maintaining high hardware efficiency.

\paragraph{Token uniqueness $\mathcal{R}^{\text{uni}}_t$}. Existing linear-attention acceleration primarily rely on chunk-wise parallel forms~\cite{kimilinear, gateddeltanet}, which strike a balance between computational efficiency and hardware parallelism by combining inter-chunk recurrence with intra-chunk parallelism. However, this leaves a challenge: under chunk-wise parallelism, the state cannot be accessed at arbitrary time steps, and the standard forward pass of linear attention permits only a single sequence of $(q,k,v)$ inputs.

As shown in Figure~\ref{fig:parallel_state_delta} (left), we first extend Gated DeltaRule for multiple queries injection. Since the injected queries are only involved in computing outputs and do not participate in state recurrence, the increase in computation is kept relatively low. After that, shown in Figure~\ref{fig:parallel_state_delta} (right), We construct pseudo queries $Q^*_t$ from input keys with shift $\delta\in[w_1, w_2)$, and perform single forward to get pseudo output matrix $\textbf{U}^*_{t,\delta}$. Then, we can re-shift the $\textbf{U}^*_{t,\delta}$ to calculate result:
\begin{equation} \label{eq:r_uni_efficient}
\begin{aligned}
    \mathcal{R}^{\text{uni}}_t=\psi(\textbf{U}^*_{t,\delta}, \textbf{U}^*_{t-1,\delta+1}).
\end{aligned}
\end{equation}

\begin{figure}[!t]
\begin{center}
\vspace{-1mm} 
\centerline{\includegraphics[width=1.0\columnwidth]{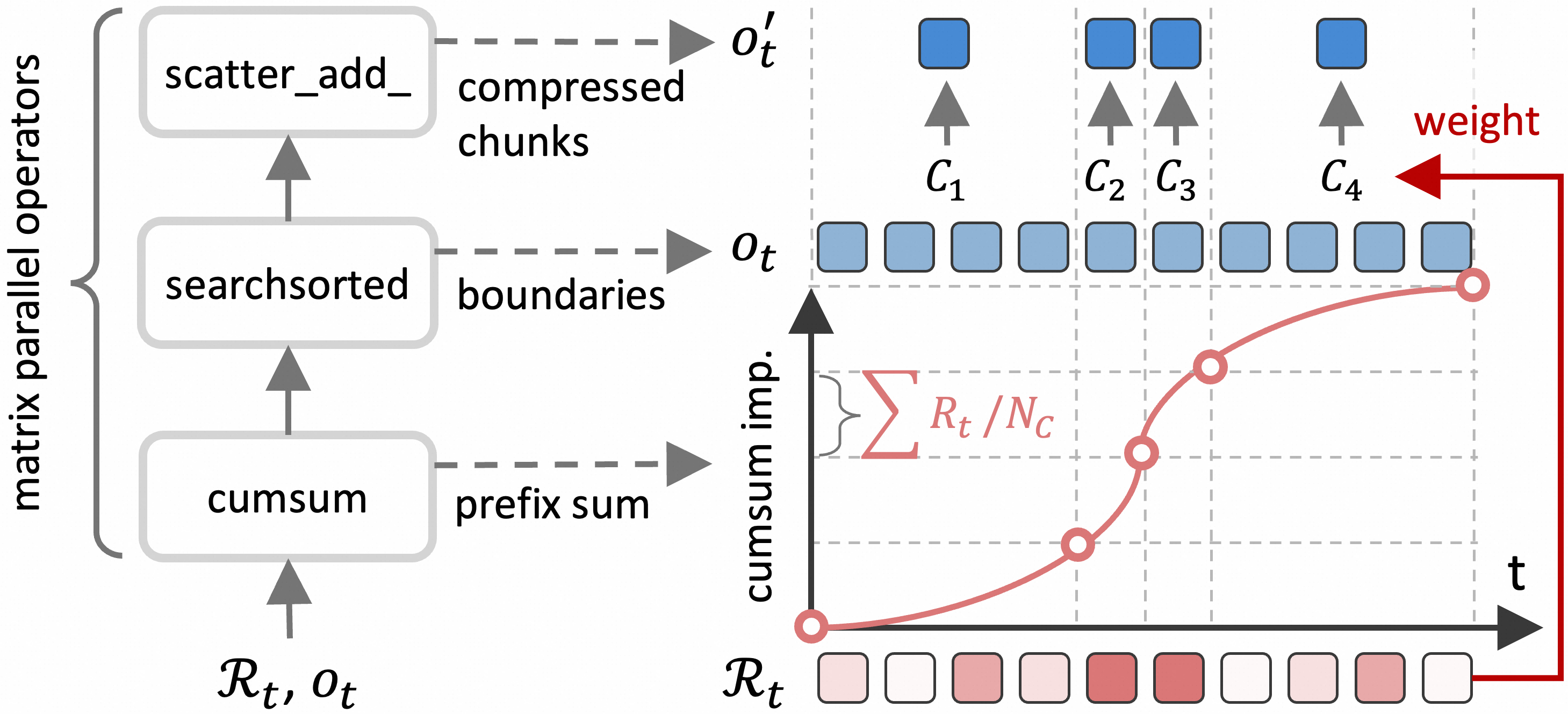}}
\caption{Efficient implementation of adaptive token merging.}
\label{fig:adaptive_token_merging}
\end{center}
\vspace{-1mm} 
\end{figure}

\paragraph{Adaptive Chunk-wise Token Merging}. Since the partitioning of the input according to importance yields a large number of chunks with varying sizes, and each chunk requires independent weight normalization for weighted aggregation, this process poses challenges to parallelization. To address this issue, as shown in Figure~\ref{fig:adaptive_token_merging}, we regard the cumulative importance curve as a continuous function and partition the importance into equal intervals according to the target number of chunks to determine the ideal chunk boundaries. We then select the actual chunk boundaries as close as possible to these ideal positions, so that the total importance of each chunk is as balanced as possible while remaining statistically unbiased. The entire process is based on parallel matrix computation, thereby maximizing GPU computational efficiency.

%% file: appendix/2_more_details.tex
\section{Evaluation Benchmarks} \label{ap_sec:more_details}
Our experimental evaluation is performed on a diverse set of benchmarks. For clarity, we provide a brief overview of these benchmarks:

\noindent\textbf{MME}~\cite{mme} evaluates the perception and cognition abilities of multimodal large language models across 14 subtasks. It adopts manually designed yes/no question pairs, where each image is paired with two questions expecting opposite yes/no answers.

\noindent\textbf{MMBench (MMB)}~\cite{mmbench} is a systematically designed objective benchmark that evaluates multimodal models through approximately 3,000 multiple-choice questions. It covers a hierarchical ability taxonomy spanning perception and reasoning, and introduces the CircularEval strategy where a model must answer correctly across all circularly shifted versions of the same question to be considered successful and produce more robust evaluation outcomes.

\noindent\textbf{GQA}~\cite{gqa} is a large-scale dataset for visual reasoning and compositional question answering, containing over 22M questions generated from Visual Genome scene graphs. It probes diverse reasoning skills such as spatial understanding, multi-step inference, and compositional generalization, and provides additional metrics for consistency, validity and plausibility.

\noindent\textbf{ScienceQA (SQA)}~\cite{sqa} is a multimodal science question answering benchmark comprising 21,208 multiple-choice questions collected from elementary and high school curricula. It spans natural science, language science, and social science, with questions accompanied by multimodal context and chain-of-thought annotations.

\noindent\textbf{TextVQA (T-VQA)}~\cite{textvqa} challenges models to recognize and comprehend textual content embedded in natural images. It contains 45,336 questions over 28,408 images sourced from Open Images, where answering correctly depends on recognizing and understanding the textual content in the scene.

\noindent\textbf{POPE}~\cite{pope} is a polling-based evaluation protocol that assesses the tendency of vision-language models to hallucinate objects. It reformulates hallucination assessment as binary yes/no questions about whether a specific object exists in the image, employing random, popular, and adversarial sampling strategies for selecting negative objects.

\noindent\textbf{SeedBench (SEED)}~\cite{seedbench} is a large-scale benchmark consisting of 19K multiple-choice questions with human-annotated answers, designed to evaluate the generative comprehension of multimodal LLMs. It spans 12 evaluation dimensions covering both image and video understanding, including spatial relation recognition and instance interaction.

\noindent\textbf{VizWiz}~\cite{vizwiz} is a VQA dataset collected from blind people who take photos with smartphones and ask spoken questions about their surroundings. It comprises over 31K visual questions and presents unique real-world challenges including poor image quality and a significant fraction of unanswerable questions.

\noindent\textbf{MMStar}~\cite{mmstar} is a vision-indispensable multimodal benchmark with 1,500 carefully curated samples across 6 capabilities and 18 detailed axes. It is specifically designed to eliminate data leakage and ensure that questions cannot be solved through text-only reasoning without actually perceiving the image.

\noindent\textbf{MMDU}~\cite{mmdu} is a multi-turn, multi-image dialogue understanding benchmark, whose samples contain up to 20 images, 27 turns and 18K multimodal tokens. Responses are scored along six dimensions, making it suitable for assessing long-form generation quality.

\noindent\textbf{V*-Bench}~\cite{vstar} contains 191 samples targeting easily overlooked small objects in high-resolution, visually crowded scenes, split into direct attribute recognition and relative spatial reasoning subsets.

%% file: appendix/3_more_detailed_analysis.tex
\section{FLOPs Analysis and Theoretical Speedup} \label{ap_sec:flops_analysis}

We analyze the computational cost of Qwen3.5's 3:1 hybrid architecture, in which every $4$ consecutive transformer layers comprise $3$ Gated DeltaNet (linear-attention) layers and $1$ \emph{gated} softmax-attention layer (with GQA and a gated $q$-projection), and derive the theoretical speedup from vision-token reduction.

\subsection{Per-Layer Computational Cost}
Consider a transformer with hidden dimension $d$, FFN intermediate dimension $d_f$, per-head dimension $d_h$, and input sequence length $N$. Each layer comprises a token mixer (attention) followed by a SwiGLU MLP. We split the per-layer FLOPs into a part linear in $N$ (projections, MLP, and the linear-attention recurrence) and a part quadratic in $N$ (softmax attention only).

\paragraph{Projections and MLP (linear in $N$).} The SwiGLU MLP contributes $6Nd\,d_f$ FLOPs. \emph{Both} layer types apply analogous QKV-style and output projections, though with different widths:
\begin{itemize}[leftmargin=*,itemsep=0pt]
    \item \emph{Softmax layer:} a \emph{gated} $q$-projection of width $2n_h d_h$, GQA $k,v$-projections of width $n_{kv} d_h$ each, and an output projection of width $n_h d_h$.
    \item \emph{Linear layer (Gated DeltaNet):} a fused QKV projection of width $2d_k + d_v$, an output-gate $z$ of width $d_v$, and an output projection of width $d$ (plus a small depthwise conv1d and tiny $a,b$ projections, all $O(Nd)$).
\end{itemize}
For the Qwen3.5 default config ($n_h d_h = d_v = d$, $n_{kv} = n_h/4$, $d_k = d/2$), the total projection output width is $\approx 4d$ for both types (softmax: $3.5d$; linear: $4d$). We therefore approximate the per-layer non-attention cost uniformly as
\begin{equation}
    \mathcal{F}_{\text{non-attn}}(N) \approx 2Nd(4d + 3d_f),
\end{equation}
where the factor of 2 accounts for fused multiply-add operations.

\paragraph{Token-mixer compute.} The attention computation itself differs between the two layer types:
\begin{itemize}[leftmargin=*]
    \item \textbf{Softmax Attention.} The $QK^\top$ product and the $\mathrm{Attn} \cdot V$ aggregation each cost $2N^2 n_h d_h$ FLOPs:
    \begin{equation} \label{eq:flops_softmax}
        \mathcal{F}_{\text{soft}}(N) = 4N^2 n_h d_h \approx 4N^2 d.
    \end{equation}
    \item \textbf{Linear Attention (Gated DeltaNet).} The recurrent update in Eq.~\ref{eq:gated_delta_update} costs $\approx c\,d_k d_v$ FLOPs per head per token; aggregated over $n_v$ value heads (keys/queries are replicated to match):
    \begin{equation} \label{eq:flops_linear}
        \mathcal{F}_{\text{lin}}(N) = c \cdot N\, n_v\, d_k d_v \approx c \cdot N d\, d_h,
    \end{equation}
    with $c \approx 8$ obtained by counting the operations in Eq.~\ref{eq:gated_delta_update} per head per token: the matrix-vector product $S_{t-1}^\top k_t$ ($2 d_k d_v$ FLOPs), the outer product $\beta_t(v_t - S_{t-1}^\top k_t)\,k_t^\top$ ($d_k d_v$), the gated combination $\alpha_t S_{t-1} + (\cdot)$ ($2 d_k d_v$), and the readout $o_t = S_t^\top q_t$ ($2 d_k d_v$), totaling $\approx 7\,d_k d_v$; we round up to $8$ to cover the qk-L2 normalization and the small $\alpha, \beta$ activations. The approximation $\mathcal{F}_{\text{lin}} \approx c N d\, d_h$ uses $n_v d_v \approx d$ and $d_k = d_v = d_h$ (Qwen3.5 default).
\end{itemize}
Thus $\mathcal{F}_{\text{soft}}$ is \emph{quadratic} in $N$ whereas $\mathcal{F}_{\text{lin}}$ is \emph{linear}, crossing over at $N = c \cdot d_h$ (e.g., $N \approx 1024$ for $d_h = 128$).

\subsection{Total FLOPs for 3:1 Hybrid Architecture}
In the 3:1 hybrid architecture (e.g., Qwen3.5, InfiniteVL), every 4 consecutive layers comprise 3 linear and 1 softmax attention layers. For a model with $\mathcal{L}$ layers, the total FLOPs are:
\begin{equation} \label{eq:flops_hybrid}
\begin{aligned}
    \mathcal{F}_{\text{hybrid}}(N) = \mathcal{L}\Bigl[
    &\underbrace{2Nd(4d + 3d_f) + \tfrac{3}{4} c N d d_h}_{\text{linear in } N} \\
    &+ \underbrace{N^2 d}_{\text{quadratic in } N}\Bigr].
\end{aligned}
\end{equation}
Letting $a = 2d(4d + 3d_f) + \tfrac{3}{4}c\,d\,d_h$ and $b = d$, this simplifies to:
\begin{equation}
    \mathcal{F}_{\text{hybrid}}(N) = \mathcal{L}\bigl(aN + bN^2\bigr).
\end{equation}

\subsection{Theoretical Speedup from Token Reduction}

Suppose vision token compression is applied at layer $l_c$, reducing $N_v$ tokens to $N_v' = rN_v$ with retention ratio $r \in (0,1)$. The effective sequence length for the remaining $\mathcal{L} - l_c$ layers becomes $N' = N_t + rN_v$. Since vision tokens dominate in VLMs ($N_v \gg N_t$), we have $N' \approx rN$, yielding a per-layer speedup:
\begin{equation} \label{eq:speedup_approx}
    \mathcal{S}(r) \approx \frac{1}{r} \cdot \frac{a + bN}{a + brN}.
\end{equation}
This reveals two regimes: $\mathcal{S}(r) \to 1/r$ when $N \ll a/b$ (linear-dominated) and $\mathcal{S}(r) \to 1/r^2$ when $N \gg a/b$ (quadratic-dominated), so the actual speedup interpolates between $1/r$ and $1/r^2$.

\paragraph{General end-to-end speedup.} Let $\rho = (\mathcal{L} - l_c)/\mathcal{L}$ denote the fraction of compressed layers, and $\gamma = bN/(a + bN)$ the relative weight of the quadratic (softmax) cost per layer. The end-to-end speedup admits a compact closed form:
\begin{equation} \label{eq:speedup_general}
    \mathcal{S}_{\text{total}}(r) = \frac{1}{(1-\rho) + \rho\, r\,\bigl[(1-\gamma) + \gamma r\bigr]},
\end{equation}
which recovers $1/r$ when $\gamma \to 0$ (linear-dominated) and $1/r^2$ when $\gamma \to 1$ (quadratic-dominated), with $\rho$ controlling how many layers benefit from compression.

As shown in Table~\ref{tab:runtime_gen}, our empirical speedups closely match Eq.~\ref{eq:speedup_general}: at $r=0.5$ with 8K vision tokens we obtain $1.91\times$ (theory $\approx 2.0\times$), and at $r=0.125$, $6.15\times$ (theory $\approx 7.0\times$).

%% file: appendix/4_additional_evaluation.tex
\section{Additional Evaluations} \label{ap_sec:additional_eval}

\paragraph{Long-form multi-turn dialogue.} To verify that compression does not harm long-form generation, we evaluate Qwen3.5-9B on MMDU~\cite{mmdu}, a multi-turn multi-image dialogue benchmark with up to 20 images, 27 turns and 18K multimodal tokens per sample. As shown in Table~\ref{tab:mmdu}, V-CoLA preserves dialogue quality across all six dimensions, remaining within 1.4 points of the full model even at a 12.5\% token budget.

\begin{table}[h]
    \centering
    \caption{MMDU results on Qwen3.5-9B under varying average vision token retention rates. All dimensions are scored by the official protocol (higher is better).}
    \label{tab:mmdu}
    \resizebox{\columnwidth}{!}{
    \begin{tabular}{lcccc}
        \toprule
        \textbf{Vision Tokens} & \textbf{100\%} & \textbf{50.0\%} & \textbf{25.0\%} & \textbf{12.5\%} \\
        \midrule
        Creativity                 & 61.0 & 61.0 & 60.6 & 60.0 \\
        Richness                   & 66.4 & 66.7 & 66.0 & 65.2 \\
        Visual Perception          & 62.2 & 62.8 & 61.7 & 61.2 \\
        Logical Coherence          & 77.5 & 77.9 & 77.3 & 76.7 \\
        Answer Accuracy            & 70.2 & 70.7 & 69.7 & 68.8 \\
        Image Relationship         & 63.1 & 64.0 & 63.0 & 62.2 \\
        \midrule
        \textbf{Overall}           & 66.6 & \textbf{67.0} & 66.0 & 65.4 \\
        \bottomrule
    \end{tabular}
    }
\end{table}

\paragraph{Fine-grained small-object recognition.} Since early exiting removes vision tokens in deep layers, we further examine tasks that rely on fine-grained visual evidence by evaluating all 191 samples of V*-Bench~\cite{vstar}, which targets easily overlooked objects in crowded high-resolution scenes. As reported in Table~\ref{tab:vstar}, exiting at layer-24 stays within 1.05 points of the full model while exiting at layer-12 loses 2.62 points, confirming the value of intermediate-layer visual access. Under a matched 50\% budget, early exiting reallocates the deep-layer token budget to shallow layers and improves overall accuracy by 2.09 points.

\begin{table}[h]
    \centering
    \caption{V*-Bench accuracy on Qwen3.5-9B. ``Pre-exit'' denotes the average vision token retention rate before the exit layer.}
    \label{tab:vstar}
    \resizebox{\columnwidth}{!}{
    \begin{tabular}{lccccc}
        \toprule
        \textbf{Setting} & \textbf{Exit} & \textbf{Pre-exit} & \textbf{Direct} & \textbf{Relative} & \textbf{Overall} \\
        \midrule
        Early exit only        & 12 & 100\%    & 70.43 & 77.63 & 73.30 \\
        Early exit only        & 24 & 100\%    & 71.30 & 80.26 & 74.87 \\
        \cellcolor{gray!15}{Original} & \cellcolor{gray!15}{32} & \cellcolor{gray!15}{100\%} & \cellcolor{gray!15}{73.04} & \cellcolor{gray!15}{80.26} & \cellcolor{gray!15}{75.92} \\
        \midrule
        V-CoLA \without exit   & 32 & 48.39\%  & 66.96 & 78.95 & 71.73 \\
        V-CoLA \with exit      & 24 & 65.22\%  & 71.30 & 77.63 & \textbf{73.82} \\
        \bottomrule
    \end{tabular}
    }
\end{table}

\paragraph{End-to-end latency.} Practical VLM inference also involves vision encoding, compression overhead and decoding, so we complement the prefill results in Table~\ref{tab:runtime_gen} with an end-to-end (E2E) profile of an INT4 TensorRT build of Qwen3.5-9B. As shown in Table~\ref{tab:e2e_latency}, V-CoLA reduces time-to-first-token (TTFT) by 38.6--64.6\% and E2E latency by up to $1.733\times$, which is particularly useful for first-response-sensitive applications. Since compression targets prefill, the attainable E2E gain scales with the prefill share of the total cost, and is thus smaller on an unoptimized PyTorch backend where prefill accounts for 29.4\% and decoding for 58.8\% of the E2E latency.

\begin{table}[h]
    \centering
    \caption{End-to-end latency (ms) of INT4 TensorRT Qwen3.5-9B on Jetson Thor with 2K visual input tokens and 32 output tokens. Overhead denotes V-CoLA's compression cost, which is already included in prefill.}
    \label{tab:e2e_latency}
    \resizebox{\columnwidth}{!}{
    \begin{tabular}{lccccccc}
        \toprule
        \textbf{Retention} & \textbf{ViT} & \textbf{Overhead} & \textbf{Prefill} & \textbf{TTFT} & \textbf{Decode} & \textbf{E2E} & \textbf{Speedup} \\
        \midrule
        \cellcolor{gray!15}{100\%} & \cellcolor{gray!15}{243.3} & \cellcolor{gray!15}{--} & \cellcolor{gray!15}{1233.0} & \cellcolor{gray!15}{1476.3} & \cellcolor{gray!15}{778.6} & \cellcolor{gray!15}{2254.9} & \cellcolor{gray!15}{$1.000\times$} \\
        50.0\%  & 243.3 & 0.47 & 662.9 & 906.2 & 778.6 & 1684.8 & $1.338\times$ \\
        25.0\%  & 243.3 & 0.48 & 402.9 & 646.2 & 778.6 & 1424.8 & $1.583\times$ \\
        12.5\%  & 243.3 & 0.46 & 279.6 & 522.9 & 778.6 & 1301.5 & $\mathbf{1.733\times}$ \\
        \bottomrule
    \end{tabular}
    }
\end{table}